\documentclass{article}

\pdfoutput=1

\usepackage{arxiv}

\usepackage[utf8]{inputenc} 
\usepackage[T1]{fontenc}    
\usepackage{hyperref}       
\usepackage{url}            
\usepackage{booktabs}       
\usepackage{amsfonts}       
\usepackage{nicefrac}       
\usepackage{microtype}      
\usepackage{lipsum}		
\usepackage{graphicx}
\usepackage{natbib}
\usepackage{doi}
\usepackage{multirow}

\title{Topic Aware Probing: From Sentence Length Prediction to Idiom Identification how reliant are Neural Language Models on Topic?}

\date{} 					

\author{ \hspace{1mm}Vasudevan Nedumpozhimana \\
	ADAPT Research Centre \\
	Technological University Dublin \\
    Dublin, Ireland\\
	\texttt{vasudevan.nedumpozhimana@tudublin.ie} \\
	\And
    John D. Kelleher \\
	ADAPT Research Centre \\
    School of Computer Science and Statistics \\
	Trinity College Dublin \\
    Dublin, Ireland \\
	\texttt{john.kelleher@tcd.ie} \\
}

\renewcommand{\shorttitle}{Topic Aware Probing}

\hypersetup{
pdftitle={Topic Aware Probing: From Sentence Length Prediction to Idiom Identification how reliant are Neural Language Models on Topic?},
pdfauthor={Vasudevan Nedumpozhimana, John D. Kelleher},
pdfkeywords={Semantics, Topic Modelling, Idiom token identification, BERT, RoBERTa},
}

\begin{document}
\maketitle

\begin{abstract}
Transformer-based Neural Language Models achieve state-of-the-art performance on various natural language processing tasks. However, an open question is the extent to which these models rely on word-order/syntactic or word co-occurrence/topic-based information when processing natural language. This work contributes to this debate by addressing the question of whether these models primarily use topic as a signal, by exploring the relationship between Transformer-based models' (BERT and RoBERTa's) performance on a range of probing tasks in English, from simple lexical tasks such as sentence length prediction to complex semantic tasks such as idiom token identification, and the sensitivity of these tasks to the topic information. To this end, we propose a novel probing method which we call \emph{topic-aware probing}. Our initial results indicate that Transformer-based models encode both topic and non-topic information in their intermediate layers, but also that the facility of these models to distinguish idiomatic usage is primarily based on their ability to identify and encode topic. Furthermore, our analysis of these models' performance on other standard probing tasks suggests that tasks that are relatively insensitive to the topic information are also tasks that are relatively difficult for these models.
\end{abstract}


\section{Introduction}

Pre-trained deep neural language models such as BERT~\citep{bert} and RoBERTa~\citep{roberta} are used to generate contextualized distributed representations (vector embeddings) of natural language text. Models based on these contextualized embeddings have achieved excellent performance across a range of NLP tasks. Consequently, what type of information is encoded in the embeddings generated by these deep neural language models is an interesting research question. 

\citet{Conneau:2018} proposed the probing methodology as a way to understand what types of information are present in an embedding. A probing task is a classification problem where a model is trained on embeddings of sentences with the goal of categorizing sentences based on a linguistic property. Examples of the types of properties that might act as the basis for the classification include the tense of the sentence, the length of the sentence, the depth of a parse tree, or the presence of particular pre-selected terms within a sentence. The probing method assumes that the success of a trained classification model (i.e., a probe) on a task indicates whether the embeddings the probe is trained on encode information relevant to the linguistic property the probe is attempting to identify.

A number of probing studies of Transformer-based language models have suggested that these models encode word-order and syntactic information in their embeddings \cite[]{raganato-tiedemann-2018-analysis, hewitt-manning-2019, clark-etal-2019-bert, reif2019visualizing, jawahar-etal-2019-bert, lin-etal-2019-open,manning2020emergent, arps-etal-2022-probing, pimentel2022architectural}. Indeed, it has been claimed that these models rediscover the classical NLP pipeline, with the earlier layers encoding syntactic information and later layers semantic \citep{tenney-etal-2019-bert} (although this claim has been questioned \citep{niu-etal-2022-bert}). Recently, a number of studies \cite[]{sinha-etal-2021-unnatural, pham-etal-2021-order, Gupta_Kvernadze_Srikumar_2021, hessel-schofield-2021-effective, sinha-etal-2021-masked} have examined the sensitivity of neural language models to word-order perturbations during pre-training, fine-tuning, and/or inference across standard benchmarks such as GLUE and PAWS and found the performance of these models is relatively insensitive to word order (although the results reported by \cite{abdou-etal-2022-word} indicate that even when trained on shuffled text neural language models do encode some word order information). One interpretation of these word-permutation results is that we need to develop more challenging benchmarks in order to really assess the linguistic abilities of modern NLP models \citep{sinha-etal-2021-masked}. A parallel interpretation is that much of the performance of these models on current benchmarks is based on shallow surface-level information such as word co-occurrence/topic, rather than syntactic information.

Given that these language models encode multiple forms of linguistic information and yet their performance on a range of tasks seems to be insensitive to word order perturbations, we are interested in examining the relative contribution of these different types of information to the improvement in NLP that Transformer-based language models have achieved in the last number of years. In particular, is this improvement primarily based on simply more effective topic modelling or is it that Transformer language models rely more on combining a variety of other linguistics signals? The concept of a topic is directly related to the idea of a coherent group of concepts, or entities in the world, that are likely to co-occur \citep{manning1999foundations} and so share a non-taxonomic semantic association \citep{kacmajor2020capturing}. Consequently, words that refer to entities/concepts that belong to the same topic are more likely to co-occur than words that refer to entities/concepts from different topics. We focus on the relative contribution of topic versus non-topic information because the topic information, understood in terms of word co-occurrence, is directly related to the masked language modelling and next-word prediction objective functions used to train many language models like BERT, RoBERTa, and GPT. However, as \cite{mickus-etal-2020-mean} pointed out, models like BERT which are also trained on next-sentence prediction objective are not purely based on distributional semantics (i.e., word co-occurrence). On the other hand, models like RoBERTa which are trained only on masked language model objective may be more focused on encoding distributional semantics. Consequently, our study of the relative contribution of topic versus non-topic information to both BERT's and RoBERTa's performance is relevant both to the current debate on the extent to which neural language models encode and use syntactic information, and also sheds light on the ongoing theoretical debate about whether these models are based on distributional semantics or not.

We approach the research question of the extent to which Transformer-based pre-trained language models rely on the topic information from two complementary directions. First, we propose a new methodology for probing which we call topic-aware probing. We also experiment using a variety of probing tasks some of which we expect to be less sensitive to topic information and others to be more sensitive. Combining the novel topic-aware probing methodology with an analysis across a range of probing tasks enables us to explore the extent to which Transformer-based models are reliant on topic versus non-topic information. We selected BERT~\citep{bert} and RoBERTa~\citep{roberta} base models 
to conduct our experiments.

Within the set of probing tasks that we examine we foreground the task of idiom token identification because the encoding of idiomatic information in neural models is relatively understudied (e.g., it is not one of the standard probing tasks proposed by \cite{Conneau:2018}), and prior research suggests that identifying idiomatic usage requires the encoding of lexical, syntactic and topic information \citep{Nedumpozhimana:2022}. An idiom is a multi-word expression with a meaning that cannot be composed of its parts \citep{Sporleder:2010}. It is hard to find a single agreed-upon definition for idioms in the literature, but they are often defined as sequences of words involving some degree of semantic idiosyncrasy or non-compositionality \citep{Fazly:2009}. Idioms appear in all languages and text genres, prototypical examples from English include expressions such as \textit{by and large} and \textit{kick the bucket} \citep{Sag:2002}. An idiomatic expression can have both idiomatic and literal meanings associated with it. \citet{Fazly:2009} highlight this aspect of idiomatic expressions and illustrate it with the expression \textit{make a face} which has an idiomatic sense in the sentence \textit{The little girl made a funny face at her mother} and has a literal sense in the sentence \textit{she made a face on the snowman using a carrot and two buttons}. Building on this distinction, \citet{Fazly:2009} defines the task of idiom token identification as deciding whether a particular usage of a given idiomatic expression is an idiomatic usage or a literal usage. While this idiom token identification task by \citet{Fazly:2009} identifies the idiomatic usage of a specific idiomatic expression in a sentence, we generalize the problem to identify the idiomatic usage within a sentence of any expression from a target category of multiword expressions. In this case, we considered the category of verb-noun idiomatic expressions from VNIC dataset \citep{Cook:2008} as our target set, but in principle, it could be any set of multiword expressions that the model is trained on\footnote{Note that in previous work \citep{Nedumpozhimana:2022} have demonstrated that it is possible to train a model to generalise from a set of known (trained on) idiomatic expressions within a category to unknown (unseen during training) idiomatic expressions from the same category.}. As shown in Table \ref{tab:sample-io}, the input of this task is a sentence that contains a usage of one of the target expressions in it (in our case, a VNIC expression) and we do not explicitly provide any information to the model regarding which idiomatic expression is present in the sentence. The same model is used to process all sentences irrespective of which expression is present in the sentence. This task of general idiom token identification is a sentence-level binary classification task and the model is required to label the input sentence as `Idiomatic' if it contains an expression from the target category that is being used idiomatically, and `Literal' if the expression is being used literally. We pay particular attention to the task of idiom token identification because a review of the literature on idiom token identification (see section~\ref{sec:ITI-lit}) reveals that this task is sensitive not only to the topic information, but also to a variety of other types of information, such as lexical and syntactic fixedness, or fluency based information. Consequently, this task provides an ideal case study to explore the relative contribution of different types of linguistics information to the performance of Transformer-based neural language models on a task.

\begin{table}[tb]
    \centering
    \caption{Sample input and corresponding target output for the general idiom token identification task}
     \begin{tabular}{@{\extracolsep{5pt}}p{10cm} c}
    \hline \hline 
    Sample input & Target output \\
    \hline 
    And we will blow our own trumpet somewhere else and sometime else. & Idiomatic \\
    The whistle has finally been blown on a controversial battle over a five-a-side football complex in Middlesbrough. & Idiomatic \\
    An angel hovered over their heads, blowing a yellow trumpet. & Literal\\
    \hline \hline
    \end{tabular}
    \label{tab:sample-io}
\end{table}

To summarise, the main contributions of this work are: (a) we present an extension to the probing method called topic-aware probing, (b) we assess the contribution of topic-based information to the performance of a Transformer-based probing on the task of general idiom token identification, and (c) more generally we explore the relationship between topic and the performance of Transformer-based neural language models across a range of probing tasks and find that tasks that are relatively insensitive to topic are also tasks that Transformer-based neural language models find relatively difficult\footnote{Our code is available at \url{https://github.com/vasudev2020/BERTAnalysis}}.

\section{Related Work}\label{sec:relatedworks}
\subsection{Transformer-based Neural Language Models}
Most of the recent neural language models are Transformer-based and many pre-trained Transformer-based language models achieve very good performance on most of the downstream natural language processing tasks. However, due to the distributed nature of the representations used by these models and the opacity of their processing of information arising from the internal complexity on Transformer neural architecture, the specific types of information these models extract from language are unclear. In response to this, there is a growing body of work focused on understanding the basis for the state-of-the-art performance of these models. For example, \cite{rogers-etal-2020-primer} surveyed over 150 papers related to the BERT model, one of the foundational Transformer-based neural language models and reviewed various kinds of information learned by BERT.

Focusing first on BERT's ability to encode syntactic information, a number of studies in the literature investigate whether BERT learns, or internally represents, any syntactic information about the input sentence. These investigations are interesting because BERT is pre-trained on a sequence of words and the pretraining objectives are not syntactic tasks. \cite{lin-etal-2019-open} showed that BERT internally represents the syntactic tree structure or an input sentence, and \cite{hewitt-manning-2019} showed that it is possible to learn transformation matrices from BERT representations which can be used to recover syntactic dependency relations within the sentences from the PennTreebank. Other studies have also shown that enough syntactic information is encoded in BERT sentence representations to allow the recovery of the parse tree structure of an input sentence \citep{vilares2020parsing,kim2020pretrained,rosa2019inducing}. However, \cite{ettinger2020bert-tacl} shows that BERT is insensitive to malformed input and argues that therefore the syntactic knowledge in BERT is either incomplete, or else BERT doesn't rely on it for solving tasks. Furthermore, \cite{glavas2021supervised} show that intermediate fine-tuning of BERT for a supervised parsing task does not improve BERT's performance, suggesting that BERT does not rely on traditional syntactic knowledge for solving tasks. The divergence between studies demonstrating BERT's ability to encode syntactic information and those that question this ability may be explained through the work of \cite{wu-etal-2020-perturbed}. \cite{wu-etal-2020-perturbed} conducted a probing experiment to assess the impact of each word on predicting other words in a masked language model. Their study found that words in the same syntactic sub-tree have a larger impact on each other. Interestingly their results also show that although BERT learns some syntactic information it is not very similar to linguistically annotated resources, and that the impact of performance on downstream NLP tasks achieved by using the syntactic structural information encoded by BERT is comparable, and even superior, to the human-designed syntactic structural information. 

These studies on BERT have been extended to other BERT-like neural language models. \cite{arps-etal-2022-probing} investigated the extent to which neural language models (namely BERT, XLNet,  RoBERTa, and DistilBERT) implicitly learn syntactic structure. They found that constituency parse trees of sentences can be extracted from distributed representations generated by these language models. Their results show that if the syntactic structure of data is correct then tree structures are extractable even if the data is semantically ill-formed. This suggests that without the help of semantic information, syntactic information can be extracted from these language models which indicates that these language models do encode syntactic information. By using a novel probing method based on the architectural bottleneck principle, \cite{pimentel2022architectural} also showed that the syntactic structure of a sentence is mostly extractable from BERT, ALBERT, and RoBERTa language models. They also point out that even though syntactic information is extractable from language models it is not clear whether this information is actually used by these models.

There are also several studies in the literature which investigate the presence of semantic information in Transformer-based neural language models. Studies conducted by \cite{ettinger2020bert-tacl} and \cite{Tenney:2019} show that BERT encodes some knowledge about semantic roles and entity relations. However, \cite{balasubramanian-2020} showed that although a BERT-based model achieves a good performance in Named Entity Recognition, there is a huge performance drop after replacing the names in the dataset, which indicates that BERT does not actually form a generic idea about named entities. By using a novel methodology to probe linguistic information for logical inference \cite{Chen_Gao_2022} observed that RoBERTa and BERT language models encode information on simple semantic phenomena rather than complex semantic phenomena.

Given that Transformer-based neural language models appear to partially encode both syntactic and semantic information about natural language text a natural question to ask is where in the Transformer architecture this information is encoded? Moreover, is the encoding of different types of information localised to specific layers in a Transformer or is it spread across multiple layers? A number of studies have investigated where the encoding of information occurs within the Transformer architecture of BERT: \cite{lin-etal-2019-open} reports word order information decreases after the 4th layer of the base-BERT model; \cite{hewitt-manning-2019} report that the reconstruction of tree depth is most successful using the middle layer embeddings of BERT (6th to 9th layers of base-BERT); \cite{goldberg-2019} show that the best subject-verb agreement is obtained by using 8th and 9th layer of the base-BERT model, and  \cite{JawaharSS19} observed that the best performances of various high-level syntactic probing tasks are achieved with middle layer embeddings of BERT. This set of results suggests that the initial layers of BERT encode information about linear word order and the later layers encode more hierarchical and syntactic information. 

There are, however, conflicting results about the presence of syntactic information in various layers of BERT. For example, \cite{Tenney:2019} and \cite{JawaharSS19} observed that the best performance on basic syntactic tasks such as POS tagging and chunking is achieved using the initial layer embeddings of BERT and good performance on high-level tasks like parsing and other semantic tasks can be achieved using embeddings from the middle layers of BERT. On the contrary, \cite{Liu:2019} observed that the best performance for both POS Tagging and chunking is obtained using middle-layer embeddings of BERT, and  \cite{Tenney:2019} observed that syntactic information is located in early and middle layers but semantic information is spread across all layers of the model. While the initial and middle layers of BERT encode syntactic and semantic information, the final layers of BERT are more task-specific, and \cite{kovaleva-2019} observed that these layers change the most while fine-tuning. This is in agreement with the observation of \cite{Liu:2019}, that overall the best performance is generally obtained using middle-layer embeddings and that embeddings from these layers are the most transferable across different tasks. 


Extending this body of work we investigate the relationship between topic and the performance of Transformer-based language models (BERT and RoBERTa) across a number of tasks. We also examine in which layers of these Transformer models the encoding of topic and non-topic information is located.

\subsection{Idiom Token Identification}\label{sec:ITI-lit}
Idioms are a sub-type of multiword expression (MWE). Other types of MWEs include compound nouns and verb particle constructions. Consequently, research on MWE is also relevant to the aspects of this research that is focused on idiom token identification. The MWE identification problem has been widely addressed within the NLP research community via the development and release of multiple shared tasks. To support research on understanding, modelling, and processing of MWEs, \cite{savary-etal-2017-parseme} introduced a shared task called PARSEME. This shared task released annotated datasets for 18 languages. \cite{ramisch-etal-2018-edition} extended this shared task to the PARSEME 1.1 task by updating the annotation methodology and releasing annotated data for 20 languages. The PARSEME shared task was further extended to the PARSEME 1.2 edition by \cite{ramisch-etal-2020-edition} in which the task involved identifying unseen MWEs and they released annotated data for 14 languages for this new task.

\cite{schneider-etal-2016-semeval} proposed a task (SemEval-2016 Task 10) which combines the labelling of multiword expressions and supersenses by the assumption that MWE and supersenses are tightly coupled. Recently \cite{tayyar-madabushi-etal-2022-semeval} proposed a task of multilingual idiomaticity detection and idiomaticity representation. The proposed idiomaticity detection is a binary task to identify whether a sentence contains an idiomatic expression with the help of two context sentences. This task has a zero-shot setting in which MWEs in the training set and test set are disjoint. It also has a One Shot setting in which the training set has one idiomatic and one non-idiomatic example of each MWE in the test set. The idiomaticity representation task is an idiomatic semantic textual similarity task, in which the semantic similarity between sentences with an idiomatic phrase, correct literal paraphrase of the idiomatic phrase, and incorrect literal paraphrase of the idiomatic phrase should be predicted.

\cite{constant-etal-2017-survey} did a detailed survey on multiword expression (MWE) processing by dividing it into two subtasks: MWE (type) discovery and MWE (token) identification. The MWE (type) discovery subtask is focused on identifying new MWEs from text and the MWE (token) identification subtask involves automatically annotating multiword expressions in running text by associating them with known multiword expression types. Our general idiom token identification task has aspects of both of these tasks in it. On the one hand, we are interested in identifying whether a particular piece of text contains a non-compositional usage from a given category of MWEs. So from this perspective, our task is similar to MWE (token) identification in that we are annotating text, although in our case the annotation is a binary label applied to the entire sentence rather than an annotation at the token level. However, because the models we train are in principle able to identify new idiomatic expressions from a given category our work also has application in the area of MWE (type) discovery, although this aspect of our work is not the primary focus in this paper (see \cite{Nedumpozhimana:2022} for work on generalising to unknown---i.e., unseen during training---expressions within a given category). 

\citet{Hashimoto:2008} reports research on idiom token identification for Japanese idioms and found that features normally used in word sense disambiguation that are defined over the context surrounding an expression worked well. Around the same time, \citet{Fazly:2009} proposed methods for idiom token identification for English based on two assumptions. First, they assumed that each idiomatic expression has a relatively fixed canonical syntactic form and that idiomatic usages of an expression tend to have this canonical syntactic form, whereas literal usages are less syntactically restricted. They also assumed that literal and idiomatic usages of an expression tend to occur with different sets of words in the surrounding context. Their results indicate that their idiom token identification model based on the syntactic form of an expression outperformed their model based on the words in the surrounding context. However, they note that this somewhat surprising result may have been affected by the fact that the definition of the typical word sets for the surrounding context of idiomatic and non-idiomatic usages used in their experiments was created using an unsupervised approach that may have resulted in noisy definitions of surrounding contexts. 

\citet{Li:2010:Cues} examined the efficacy of feature sets based on global lexical context, discourse cohesion and local lexical features, such as cue words, for idiom token identification. They found that features based on global lexical context and discourse cohesion were the most effective. \citet{Li:2010:Gauss} confirmed the efficacy of discourse cohesion features for idiom token identification. Following this theme of contextual approaches to idiom token identification, \citet{Feldman:2013} and \citet{Peng:2014} explored topic features for the idiom token identification problem. \citet{Feldman:2013} and \citet{Peng:2014} based their work on the assumption that idiomatic usages will be semantically distant from the topics of the discourses in which they are present and so a candidate expression should be identified as an idiomatic usage if it is a semantic outlier with respect to the surrounding context. 

\citet{Salton:2016} demonstrated the viability of building an idiom token identification model using a distributed sentence representation, specifically the sentence embeddings generated by Skip-Thought \cite[]{Skip-Thought}. 
They proposed a 
model for both per-expression and general idiom token identification problems (i.e., developing a single idiom token identification model that works across multiple expressions from a given category of MWE). 
Unlike previous work which required separate contextual/topic-based models for each expression being assessed, a distinctive aspect of the work by \citet{Salton:2016} is that, unlike previous work which required separate contextual/topic-based models for each expression being assessed, their approach used a single model across all expressions within a category. Furthermore, their model only required the distributed embedding of the sentence the expression occurs within, and did not need access to the surrounding context of the sentence. 
A natural question arising from the results reported by \citet{Salton:2016} is what are the kinds of information that the distributed representation of a sentence encodes which are so useful for idiom token identification? For example, these embeddings may be capturing syntactic or lexical fixedness features, similar to those proposed by \citet{Fazly:2009}, or be efficiently encoding some form of topic-based signal (efficient both in the sense of only requiring a small sample of text---i.e., the sentence---to pick up the relevant context, and also in terms of being able to do this across multiple expressions with, presumably, variation in the topic signals associated with each expression). \cite{hashempour-villavicencio-2020-leveraging} also performed an idiom token identification experiment and found that contextual word embeddings, such as those generated by BERT, outperform non-contextual word embeddings. However, probing experiments by \cite{garcia-etal-2021-probing} on contextualised vector space models like ELMo and BERT concluded that idiomatic usage is not yet accurately represented by these contextualised models. Extending the work on using neural sentence embeddings for idiom token identification, \cite{nedumpozhimana2021finding} showed that more recent contextual distributed representations such as those generated by BERT models encode idiomatic information and their result suggest that a topic signal might be the key information encoded in the BERT representation.

More recently, \citet{Nedumpozhimana:2022} report experiments using the game theory concept of Shapley Values to analyse the type of information that idiom token identification models based on BERT find useful. They first report expression-wise experiments using Shapley Value analysis–that analysed the relative contribution of different expressions to the generalisation ability of an idiom token identification model. Then they used the results of these experiments together with an expression-wise analysis of the association between different expressions and different linguistic phenomena (syntactic and/or lexical fixedness, topic, and so on) to assess which types of linguistic information are more useful for idiom token identification. They find that a combination of idiom-intrinsic and topic-based features are useful for achieving generalizability, and argue that their results point to BERT encoding different types of linguistic information, including topic, lexical and syntactic information. Prompted by these findings, within our examination of the role of the topic information in the distributed representations generated by Transformer based neural language models, we put a particular focus on the extent to which Transformer-based models (BERT and RoBERTa) rely on the topic information to identify idiomatic usage. Accordingly building on previous work that performed expression-wise analysis here we directly analyse the contribution of the topic information in general by assessing the ability of general idiom token identification models trained on sentences from one topic to generalize to sentences from other topics.

\section{Topic-Aware Probing}\label{sec:topic-aware-prob}
Our topic-aware probing method is designed to investigate the role of topic signals in a probing task. The basic idea is to train the probing model on samples from a particular topic and then test it on samples from the topic the training data were sampled from and separately on samples from other topics. We then analyse the difference in performance on the samples from the topic seen during training and the samples from unseen topics. A large difference in performance would indicate that the topic information is an important factor in determining the performance of a model on a task. 

\begin{figure}
    \centering
    \includegraphics[width=10cm]{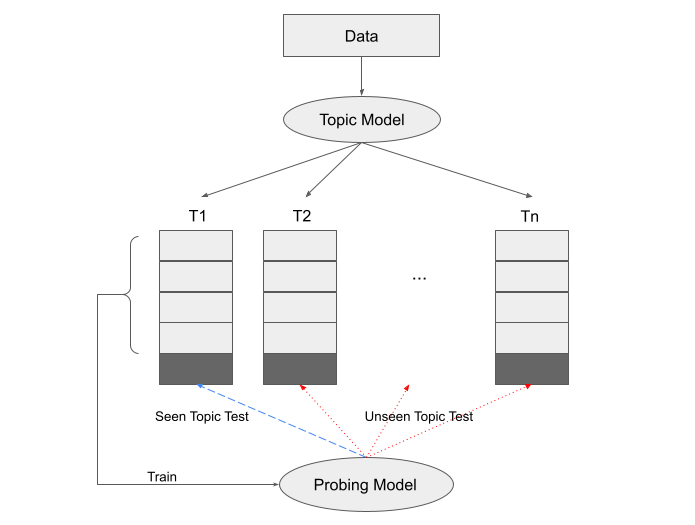}
    \vspace{0.5cm}
    \caption{Topic-Aware probing method}
    \label{fig:topic-aware-probe}
\end{figure}

The first step in topic-aware probing is to partition the dataset into $n$  different topics using a topic model. Next, we split the set of samples in each topic into $k$ folds, for cross-fold validation. We then iterate through the topics and for each fold $i$ in a topic we:
\begin{enumerate}
    \item train the probe (e.g., the general idiom token identification model) using the data from the other folds in the topic; 
    \item evaluate the probe on the $i^{th}$ fold in the topic and record the performance of the probe as a \emph{seen topic score}; 
    \item iterate through the other $n{-}1$ topics in the topic model, evaluate the probe on the corresponding $i^{th}$ fold in these other topics, and record the performance of the probe on each of these folds as an \emph{unseen topic score}.
\end{enumerate}
At the end of this process, for each of the $n$ topics we have calculated $k$ seen topic scores and $k \times (n-1)$ unseen topic scores. We then calculate for each topic the average seen topic score and the average unseen topic score. Figure \ref{fig:topic-aware-probe} illustrates the topic-aware probing method.

If the topic signal is important in terms of helping the probe to predict the label for the task then the performance of the probe model should be significantly better on samples from seen topics compared to unseen topics (i.e., the seen topic scores should be higher than the unseen topic scores). On the other hand, if we do not observe a significant difference in performance between samples from seen topics compared to unseen topics this would indicate that topic is not an important signal in terms of the probing task. 

In our experiments, we use Latent Semantic Indexing (LSI) (also known as Latent Semantic Analysis) for topic modelling: LSI is an unsupervised topic-modelling approach based on the distributional hypothesis \cite[]{LSI}. We use LSI because it captures word co-occurrence \cite[]{manning1999foundations,eisenstein2019introduction} which is essentially the topic signal that we are assessing in this experiment. 

For cross-fold validation, we use 5 folds, and in order to maintain the label distribution within a topic across the 5 folds we use stratified sampling to create the folds. Note that some of the topics identified by the topic model can have less than 5 samples from some label categories and so for these topics, we would not be able to split the samples into 5 stratified folds. One option for dealing with these small topics is to discard them. This, however, would result in some labelled examples being discarded, something that may be undesirable if we are working with a small dataset. Consequently, we consider any topics with less than 5 samples from some label as a tail topic, and we iteratively merge the tail topics with other topics (preferably with some other tail topic) to reduce the number of tail topics. The iterative procedure for merging tail topics is as follows: if there is only one tail topic then we randomly select one of the non-tail topics and merge the tail topic with the selected non-tail topic, and if there is more than one tail topic then we randomly select two of the tail topics and merge them. This process of tail topic merging continues until there is no tail topic left. This tail-reduction process may reduce the semantic coherence within some topics because it involves merging unrelated topics into a single topic. Furthermore, this dilution in the topic signal may result in the topic-aware probing method underestimating the importance of the topic signal on a given task (i.e., it may increase the likelihood of a Type II error—false-negative—in a topic-aware probing experiment). In other words, if topic merging has an effect on an experimental analysis the effect is to reduce the sensitivity of the method to the topic signal by reducing the difference between seen and unseen scores. Consequently, in situations where we do see a difference between seen and unseen scores the merging of topics will only have weakened this difference, and not caused it. So when we see a difference, topic merging won’t be the cause of the difference. The more problematic case is where we don’t see a difference between seen and unseen scores. In this case, a topic difference may in fact exist but the merging may have diluted it. Fortunately, however, the problem of tail topics typically only arises for runs of the topic modelling process where we extract a large number of topics from a relatively small dataset. The number of tail topics obtained from different topic models on different datasets are shown in the coming sections in Table \ref{tab:no-topics} and Table \ref{tab:no-topics-other-probes}. Consequently, the topic merging process does not affect all the runs of a topic modelling process, and so one way to mitigate the effect of topic merging is to report the average seen versus unseen difference across multiple runs of a topic model with different numbers of topics identified in each run. This is one of the reasons why in our experiments we report the average difference between seen and unseen topic scores across 10 topic models with the number of topics ranging from 5 to 50. The other reason is that by reporting average differences across multiple topic models we reduce the sensitivity of the analysis to the number of topics chosen by a single topic modeling process. Once the tail topics are merged we split samples from each topic into 5 stratified folds.

Multilayered perceptron (MLP) models with one hidden layer are one of the standard models used in the probing literature \citep{Conneau:2018}. Furthermore, probing experiments have demonstrated that dense embeddings, such as BERT embeddings, can encode information in a distributed manner (e.g., in the embedding norm \citep{klubickakelleher2022}) and so using a model type that is able to integrate information from across an embedding (such as an MLP) allows the probe to utilize this distributed information. Therefore we use an MLP model, with one hidden layer using ReLU as the hidden layer activation function, as the probing model for predicting the label from the distributed representation of a sample sentence. 

\section{Experimentation}\label{sec:experimentation}
In order to confirm that topic-aware probing functions as expected, we first apply it to a probing task that we expect will not be sensitive to the topic information, namely bigram shift---a probing task introduced by \citet{Conneau:2018}. The bigram shift task is to predict whether any two consecutive words within a sentence have been swapped. We do not expect the topic signal to be a useful information source for this task because swapping two consecutive words in a sentence will not change the topic of the sentence (at least not from the perspective of a word co-occurrence-based topic model because the sentence will contain the same set of words after the swapping as the original sentence). Consequently, the bigram shift task will enable us to validate topic-aware probing. Our expectation is that because the bigram shift task is not sensitive to topic information we should observe similar seen and unseen topic scores if the methodology is working as expected. 

Having checked the topic-aware probing methodology works as expected for the bigram shift task we switch our focus to investigate the role of the topic as a signal in BERT-based and RoBERTa-based general idiom token identification. To do this we first confirm that general idiom token identification is sensitive to topic signals by using topic-aware probing, and then isolate the contribution of the topic signal to the performance of BERT and RoBERTa on this task by comparing the performance of BERT and RoBERTa to a (primarily) topic-based embedding model. We use GloVe embeddings to act as this topic baseline because GloVe embeddings are trained on the nonzero elements in a word-word co-occurrence matrix \cite[]{Pennington:2014}, and so the primary information captured by GloVe embeddings directly relates to the concept of the topic we are examining here. Furthermore, as we shall see in Section \ref{sec:results}, our analysis of the results from applying topic-aware probing to the bigram shift task not only confirms that the topic-aware probing method is functioning as expected but also supports the assumption that GloVe embeddings primarily encode topic (word co-occurrence) information.

\subsection{Data Preparation} \label{subsec:dataprep}
For the experiments on the bigram shift task, we used 119,998 English sentences from the established bigram shift dataset\footnote{\href{https://github.com/facebookresearch/SentEval/tree/master/data/probing}{https://github.com/facebookresearch/SentEval/tree/master/data/probing}}. The bigram shift dataset labels original sentences as  `Original' and inverted sentence as `Inverted'. The dataset contains 59,999 `Inverted' sentences and 59,999 `Original' sentences.

The experiments on the general idiom token identification task are based on the VNIC dataset \cite[]{Cook:2008}. The VNIC dataset is a set of 2,979 English sentences with each sentence containing an instance of one of 53 idiomatic expressions. An idiomatic expression can be used either in an idiomatic or literal sense. The VNIC dataset contains manually annotated labels, where every sentence is marked as ‘Idiomatic usage’, ‘Literal usage’ or ‘Unknown’. If the idiomatic expression in the sentence is used idiomatically then the sentence will be labelled as `Idiomatic usage', if the idiomatic expression is used literally then the sentence will be labelled as `Literal usage', and if the usage is ambiguous for human annotator then the sentence will be marked as `Unknown'. The sentences with the `Unknown' label are either idiomatic or literal samples but the human annotator was unable to decide whether it is idiomatic or not. Unfortunately, from a practical point of view, the `Unknown' sentences without a manually annotated label (idiom or literal) are impossible to use for training or for evaluation in a supervised learning setup. Therefore we removed all such sentences from our dataset for our experimentation. The removal of these sentences from the dataset does simplify the task of general idiom token identification with respect to the task that humans processing language face. However, the focus of our analysis is on understanding what linguistic information (topic versus non-topic) Transformer-based pre-trained language models like BERT and RoBERTa use when they are processing language, and so removing these `Unknown' sentences may lead to a cleaner signal within the analysis of the performance of these models on the task. This filtering of the VNIC dataset left 2,566 samples of which 2,016 were idiomatic usages and 550 were literal usages across the 53 expressions. When preparing the training and test sets we split the data by sentence rather than by expression, and so an idiomatic expression may appear in sentences in the training set and the test set\footnote{There are multiple samples for each of the 53 idiomatic expressions in the general idiom token identification dataset---some marked as `Idiomatic usage', others as `Literal usage'---and these samples may be distributed across multiple topics. Consequently, some topics may include both `Idiomatic usage' and `Literal usage' samples for some expressions, or only samples of one type for an expression, or no samples for a given expression. Consequently, the seen topic versus unseen topic distinction is different to the seen expression versus unseen expression distinction examined by \citet{Salton:2016} (which is not our focus in this paper).}.

\subsection{Representations}\label{subsec:representation}
For our probing experiments, we generated a distributed representation of each sample sentence by using a pre-trained BERT model (\textit{bert-base-uncased}) and a pre-trained RoBERTa model (\textit{roberta-base}). Both these models are based on the Transformer encoder architecture with 12 layers, 768 hidden dimensions, and 12 attention heads. \citet{JawaharSS19} observed that different layers of the BERT architecture capture different kinds of information. So we generated different embeddings from each of the 12 different layers of both BERT and RoBERTa architectures. Our models have 12 layers and therefore we generated 12 different BERT and RoBERTa embeddings for each sample sentence. In each layer, we generated an aggregate distributed representation of the sentence by averaging the distributed representations of each token in the sentence. There are a number of ways that we could have generated a sentence embedding from BERT and RoBERTa, for example, we could have used the CLS token. However, \cite{mosbach-etal-2020-interplay} suggests that, for probing tasks, the average of the embeddings of the token in a sentence is a better sentence-level representation than the embedding of the CLS token. Also, as we explain below, in our experiments we use GloVe as a baseline topic-based distributed representation, and using the average token embedding for BERT and RoBERTa makes the process we use to generate BERT and RoBERTa sentence representations more consistent with the process we use to generate GloVe sentence representations, which is the average of the GloVe representation of all words in the sentence.

GloVe is a distributed representation based on a word-to-word co-occurrence matrix \cite[]{Pennington:2014}. This approach is very similar to the topic modelling approaches, particularly to Latent Semantic Indexing, the approach we are using for our probing experiment. So we assume that GloVe embeddings primarily capture topic information of words from a large corpus. To generate the GloVe representation of a sample sentence we averaged the GloVe representations of each word in it. Therefore the GloVe embeddings used in our experiment likely neglect the syntactic structure of the sentence, and this is a deliberate choice as part of our methodology so that GloVe can be used as a metric for the expected performance of an embedding that primarily encodes topic on a probing task. We consider GloVe distributed representations as one of the baseline representations for topic-aware probing. The comparison against the GloVe baseline enables us to estimate the amount of topic and non-topic signal BERT and RoBERTa encode.

\citet{HewittL19} warned that the probe itself can learn the task without using the information in the sentence representation especially when the probing model is powerful enough to capture the task objective. To control for this possible confounding factor we also train a baseline probe model for each task on random vector representations to measure the performance due to the power of the probing model. The random vector representations are created by randomly generating a 768-dimensional vector for each of the input sentences (we use 768-dimensional vectors so that the random vectors have the same dimension as the BERT and RoBERTa embeddings). 

\subsection{Experimental Design}\label{subsec:probing-experiment}
In the topic-aware probing with LSI topic modelling, we have to specify the number of topics. If the number of topics is kept small then each topic will be more generic and this may reduce the power of the topic-aware probing, although this may be mitigated by the fact that a small number of topics will also result in larger training and test sets. Conversely, if the number of topics is too large then each topic will be very specific but the training and testing sample size will be reduced and this may result in the underperformance of the probe. This will be reflected in the seen scores and unseen scores and therefore in our analysis. 

In order to control for the confounding effects of the number and size of topics on our analysis, for both probing tasks (bigram shift and general idiom token identification) we repeat our topic-aware probing experiment 10 times while varying the number of topics from 5 to 50 in increments of 5. Consequently, each iteration of an experiment uses a different topic model in the topic-aware probing as the basis for the experiment. Note that as the number of topics approaches 50, the chance that the initial topic model in the topic-aware probing will contain topics with $<5$ samples with some label (i.e., tail topics) increases. The actual number of tail topics from each of the 10 topic models on both the Bigram shift and VNIC datasets is shown in Table \ref{tab:no-topics}. In such cases, the actual number of topics used for probing will be less than the specified number of topics after the tail reduction (as discussed in Section \ref{sec:topic-aware-prob}).

 \begin{table}
    \centering
   \caption{Number of tail topics from 10 topic models on Bigram shift and VNIC dataset}
     \begin{tabular}{@{\extracolsep{1pt}}l c c c c c c c c c c c}
     \hline \hline 
    Max number of topics & 5 & 10 & 15 & 20 & 25 & 30 & 35 & 40 & 45 & 50 \\\hline	
    Bshift & 0 & 0 & 0 & 0 & 0 & 0 & 1 & 0 & 0 & 0 \\
    VNIC & 0 & 2 & 1 & 4 & 6 & 11 & 13 & 18 & 20 & 18 \\\hline \hline
     \end{tabular}
   \label{tab:no-topics}
 \end{table}

To apply our topic-aware probing we first divide the dataset into different partitions with different topics. The topic model can divide samples with the same class label into the same partition (or same set of partitions) and in such cases, we can see that the topic model itself will internally do the classification task. Similarly, in the case of the general idiom token identification dataset, the topic model can divide samples with the same idiomatic expression into the same partition (or the same set of partitions). We initially checked for such an interaction between topics, labels, and expressions by investigating how evenly the labels and expressions are distributed across different topics. For that, we calculated the \textit{mean normalized entropy} of the distributions of each class label and each expression. To calculate the \textit{mean normalized entropy} of a distribution we first calculated the normalized entropy of the distribution across different topics in each of the 10 topic models and then averaged it. The normalized entropy of a probability distribution is the entropy of the distribution normalized with the maximum possible entropy\footnote{The normalized entropy of a label (or an expression) distribution $p$ across a set of $n$ topics is $ -\sum_{i=1}^{n} \frac{p(x_i)log_b(p(x_i))}{log_b(n)}$. For example the normalized entropy of distribution: $(1,0,0)$ will be $0$, $(\frac{1}{3},\frac{1}{3},\frac{1}{3})$ will be $1$, and $(\frac{1}{2},\frac{1}{2},0)$ will be $0.63$}. An interesting property of normalized entropy is that the value will be in the range of 0 to 1 and it will be independent of the base of the logarithm. If the distribution is uniformly distributed then we will get the maximum normalized entropy 1 and if the distribution is highly skewed (i.e., samples with the same class label are in the same partition or samples with the same expressions are in the same partition) then we will get the minimum entropy 0. In the VNIC dataset, the \textit{mean normalised entropy} across topics for the `Idiomatic usage' label was $0.85$ bits and for the `Literal usage' label was $0.81$ bits. In the bigram shift dataset the \textit{mean normalised entropy} across topics for the `Original' label and the `Inverted' label was $0.86$ bits. These large entropy values indicate that on average both labels in both datasets are relatively evenly distributed across topics (i.e., the topic model process was neither doing general idiom token identification nor bigram shift classification). The average \textit{mean normalised entropy} of distributions from different idiomatic expressions in the VNIC dataset is $0.39$ bits. This entropy is lower than that of the label distributions. Also, we note that there are a few expressions that have a \textit{mean normalised entropy} lower than $0.2$ bits and for one expression the \textit{mean normalised entropy} is less than $0.1$. This suggests that for these expressions with low entropy, most of the samples are grouped into the same topic. One intuitive explanation for this is that for each expression the literal instances tend to cluster within a topic and the idiomatic instances tend to be distributed across topics. To test this intuition we calculated the \textit{mean normalised entropy} of the distribution of the expressions across topics when we consider only the literal instances of the expression, this entropy was found to be $0.3583$, and when we only consider the idiomatic instances the entropy was $0.4098$. The fact that in general, the distribution of an expression’s idiomatic instances across topics has a higher entropy than the literal instances suggests that our intuition is correct.

\begin{figure}
    \centering
    \includegraphics[width=10cm]{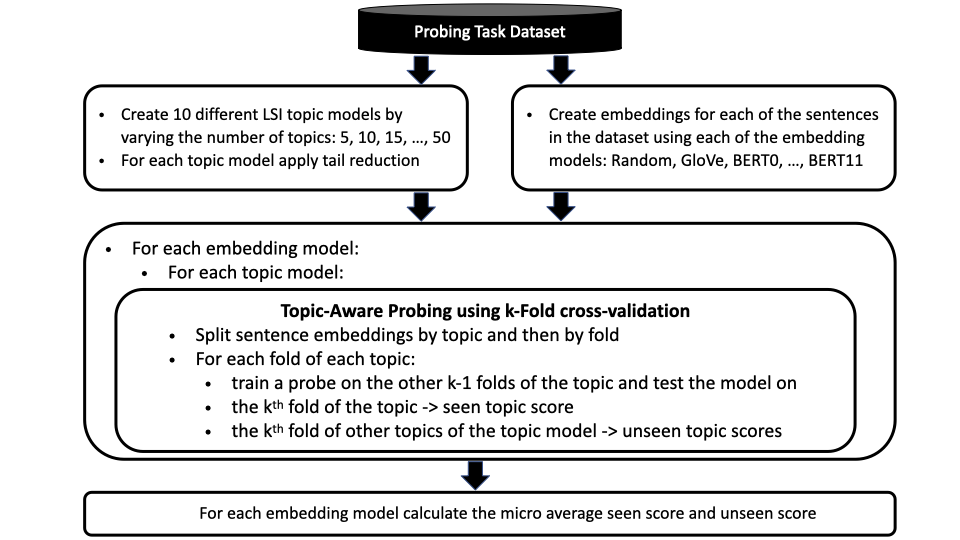}
    \vspace{0.5cm}
    \caption{Experimental design}
    \label{fig:experimental-design}
\end{figure}

For each topic-aware probing (i.e., for each combination of task plus embedding) with $n$ topics, we get $n$ seen and $n$ unseen scores (each being an average across 5 folds). We have averaged (a micro average) these seen scores and unseen scores from all topic models to calculate an aggregate seen score and unseen score. We then calculate the average difference by taking the difference between these two averages\footnote{Note that the difference between the averages is the same as the average of the differences.}. If the average difference is positive and not a negligibly small value this is an indication that the topic signal contributes to the performance of the probing model on the task. Our experimental design is illustrated in Figure \ref{fig:experimental-design}.

The label distribution in the VNIC dataset for general idiom token identification is somewhat imbalanced. Most imbalanced datasets contain more negative samples and fewer positive samples. However, somewhat unusually in this case, there are more positive samples (idiomatic usage) than negative samples (2566 idiomatic and 550 literal usages). 
\cite{savary-et-al:2019} proposed heuristics to automatically identify sample sentences that might contain literal occurrences of MWE and this approach can be considered for balancing the general idiom token identification dataset. However, using this approach would still require manual verification and therefore in our experimentation we used the VNIC dataset as it is and selected an evaluation metric that accounts for imbalanced distributions. Accuracy is a popular and intuitive evaluation metric, but it is not suitable for an imbalanced dataset: if 99 samples out of 100 are idiomatic, a model could report high accuracy by blindly labelling every sample as positive. The F1 score and Area Under Precision Recall curve (AUC-PR) are both suitable for the standard imbalanced scenario where the positive class is the minority, this is because they both focus on the identification of positive samples. However, because they both exclusively focus on the performance of the positive class they are not suitable when the positive class is the majority. In this context, although we could use the F1 scores as our evaluation metric by treating the literal class as the positive class, doing this would essentially change the task to literal token identification and so would add an extra layer of complexity to the interpretation of the results in terms of idiom identification. An alternative is to use other metrics that are suitable for imbalanced datasets that consider both the positive and the negative classes. The most suitable evaluation metrics for an imbalanced dataset with a majority of positive instances are the AUC ROC and Mathew Correlation Coefficient (MCC). An empirical comparative study by \citet{halimu} showed that both AUC ROC and MCC are statistically consistent with each other, however, AUC ROC is more discriminating than MCC. Therefore we selected the AUC ROC as the most suitable metric for these experiments.

For this experiment, we used the gensim\footnote{https://radimrehurek.com/gensim/} library implementation of Latent Semantic Indexing with default parameters for training the topic model. We used the bigram phrase model, tf-idf model, and data lemmatization from gensim library to prepare the corpus for training the LSI model. For training the MLP probing model we used the scikit-learn \cite[]{scikit-learn} implementation of MLP with default parameters.

\section{Results and Discussion}\label{sec:results}
We report results obtained from our probing experiments: micro averaged seen and unseen AUC-ROC scores, and the differences between seen and unseen scores for each representation in Table \ref{tab:results}.

 \begin{table}
   \caption{Average Seen and Unseen AUC ROC scores and their differences along with standard deviations for different embeddings on the Bigram Shift Probing task and the General Idiom Token Identification task}
     \begin{tabular}{@{\extracolsep{1pt}}p{0.55in} c c c c c c}
     \hline \hline 
     \multirow{3}{*}{} & \multicolumn{3}{c}{Bigram Shift} & \multicolumn{3}{c}{General Idiom Token Identification} \\
     \cline{2-4}     
     \cline{5-7}
     & Seen  & Unseen  & Difference & Seen & Unseen & Difference \\
    & Mean(Stdev) & Mean(Stdev) & Mean(Stdev) & Mean(Stdev) & Mean(Stdev) & Mean(Stdev) \\     \hline
Rand & 0.4983(0.0216) & 0.4996(0.0055) & -0.0013(0.0223) & 0.5050(0.1215) & 0.4966(0.0217) & 0.0083(0.1196) \\
GloVe&0.5088(0.0182)&0.5013(0.0033) & 0.0075(0.0184) &0.7747(0.1231)&0.6571(0.0664)&0.1177(0.1227)\\\hline	
BERT0	&	0.5333(0.0351)	&	0.5217(0.0106)	&	0.0116(0.0321)	&	0.7896(0.1133)	&	0.6917(0.0582)	&	0.0979(0.1157)	\\
BERT1	&	0.6926(0.0535) &	0.6671(0.0402)	&	0.0255(0.0340)	&	0.8026(0.1163)	&	0.6984(0.0610)	&	0.1041(0.1172)	\\
BERT2	&	0.8620(0.0413) &	0.8444(0.0378)	&	0.0176(0.0244)	&	0.8150(0.1218)	&	0.7009(0.0655)	&	0.1141(0.1211)	\\
BERT3	&	0.8982(0.0325)	&	0.8848(0.0283)	&	0.0134(0.0190)	&	0.8372(0.1145)	&	0.7270(0.0632)	&	0.1102(0.1111)	\\
BERT4	&	0.9155(0.0295)	&	0.9041(0.0248)	&	0.0114(0.0186)	&	0.8468(0.1131)	&	0.7492(0.0605)	&	0.0976(0.1144)	\\
BERT5	&	0.9276(0.0271)	&	0.9187(0.0237)	&	0.0089(0.0172)	&	0.8536(0.1055)	&	0.7605(0.0575)	&	0.0931(0.1059)	\\
BERT6	&	0.9367(0.0256)	&	0.9275(0.0226)	&	0.0092(0.0166)	&	0.8560(0.1112)	&	0.7651(0.0565)	&	0.0909(0.1101)	\\
BERT7	&	0.9437(0.0205)	&	0.9360(0.0198)	&	0.0077(0.0143)	&	0.8610(0.1090)	&	0.7714(0.0564)	&	0.0895(0.1118)	\\
BERT8	&	0.9447(0.0165)	&	0.9361(0.0169)	&	0.0086(0.0147)	&	0.8593(0.1096)	&	0.7711(0.0570)	&	0.0881(0.1141)	\\
BERT9	&	0.9352(0.0204)	&	0.9258(0.0200)	&	0.0094(0.0155)	&	0.8569(0.1156)	&	0.7605(0.0583)	&	0.0964(0.1208)	\\
BERT10	&	0.9301(0.0226)	&	0.9204(0.0205)	&	0.0097(0.0166)	&	0.8521(0.1157)	&	0.7537(0.0571)	&	0.0984(0.1167)	\\
BERT11	&	0.9215(0.0201)	&	0.9103(0.0181)	&	0.0112(0.0167)	&	0.8414(0.1309)	&	0.7289(0.0555)	&	0.1125(0.1306)	\\\hline
RoBERTa0	&	0.7424(0.0648)	&	0.7194(0.0482)	&	0.0230(0.0367)	&	0.8141(0.1341)	&	0.6807(0.0656)	&	0.1334(0.1331)	\\
RoBERTa1	&	0.8380(0.0623)	&	0.8198(0.0512)	&	0.0182(0.0289)	&	0.8366(0.1311)	&	0.7130(0.0718)	&	0.1236(0.1308)	\\
RoBERTa2	&	0.8903(0.0503)	&	0.8762(0.0435)	&	0.0141(0.0247)	&	0.8574(0.1231)	&	0.7405(0.0668)	&	0.1169(0.1204)	\\
RoBERTa3	&	0.9135(0.0364)	&	0.8999(0.0340)	&	0.0136(0.0256)	&	0.8696(0.1168)	&	0.7504(0.0643)	&	0.1192(0.1182)	\\
RoBERTa4	&	0.9222(0.0335)	&	0.9112(0.0294)	&	0.0111(0.0234)	&	0.8775(0.1100)	&	0.7641(0.0631)	&	0.1134(0.1141)	\\
RoBERTa5	&	0.9252(0.0326)	&	0.9148(0.0276)	&	0.0104(0.0220)	&	0.8812(0.1069)	&	0.7746(0.0618)	&	0.1067(0.1082)	\\
RoBERTa6	&	0.9261(0.0278)	&	0.9144(0.0257)	&	0.0117(0.0201)	&	0.8809(0.1106)	&	0.7738(0.0615)	&	0.1071(0.1095)	\\
RoBERTa7	&	0.9304(0.0293)	&	0.9199(0.0247)	&	0.0105(0.0192)	&	0.8877(0.1060)	&	0.7815(0.0580)	&	0.1062(0.1051)	\\
RoBERTa8	&	0.9355(0.0317)	&	0.9261(0.0250)	&	0.0093(0.0193)	&	0.8876(0.1117)	&	0.7801(0.0577)	&	0.1075(0.1142)	\\
RoBERTa9	&	0.9390(0.0313)	&	0.9306(0.0244)	&	0.0084(0.0196)	&	0.8930(0.1062)	&	0.7882(0.0569)	&	0.1048(0.1098)	\\
RoBERTa10	&	0.9369(0.0354)	&	0.9286(0.0255)	&	0.0083(0.0246)	&	0.8902(0.1105)	&	0.7805(0.0592)	&	0.1097(0.1165)	\\
RoBERTa11	&	0.9291(0.0312)	&	0.9180(0.0240)	&	0.0111(0.0242)	&	0.8772(0.1133)	&	0.7523(0.0634)	&	0.1249(0.1177)	\\	
     \hline \hline 
     \end{tabular}
   \label{tab:results}
 \end{table}

\begin{figure}
    \centering
    \includegraphics[height=4.7cm]{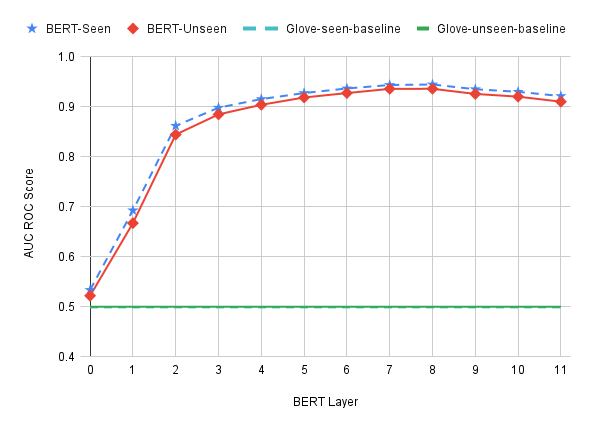}
    \includegraphics[height=4.7cm]{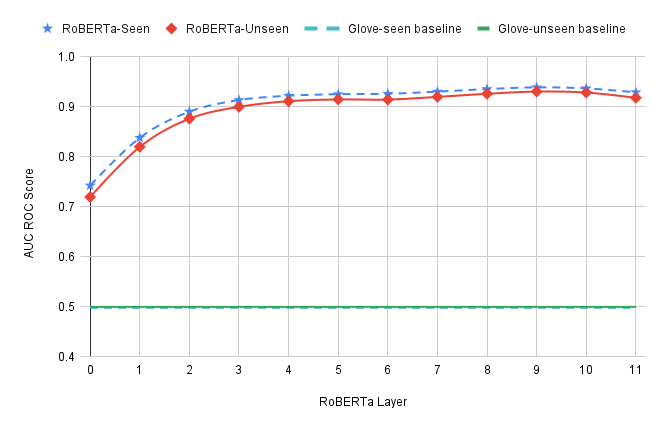}
    \vspace{0.5cm}
    \caption{Seen and Unseen AUC ROC scores from GloVe and different layers of BERT and RoBERTa on the Bigram Shift Task}
    \label{fig:bshift-scores-layers}
\end{figure}

Focusing first on the bigram shift task we observe very small differences in performance between seen AUC scores and unseen AUC scores using any of the representations (GloVe, BERT, or RoBERTa) with a maximum 2.55\% difference, and in most of the cases less than 1\% difference. This is evident in Figure \ref{fig:bshift-scores-layers} where the GloVe seen and unseen scores are plotted on top of each other as are the BERT and RoBERTa seen and unseen scores across all the layers. This is an expected result (see Section \ref{sec:experimentation}) and one that we take to indicate that topic-aware probing works as expected. Furthermore, this result also indicates that the bigram shift task is not sensitive to a topic signal.

Given that the bigram shift task does not appear to be sensitive to the topic signal it is interesting to observe that GloVe embeddings and random embeddings give the same performance on this task. If GloVe encodes non-topic signals along with topic signals then we would expect that this information would help the GloVe-based probing model to achieve better performance (as compared to a random embedding) on a non-topic sensitive task such as bigram shift. But in this case, GloVe has the same performance as the random baseline, which suggests that GloVe embeddings do not encode non-topic signals. This is in line with our assumption that GloVe primarily captures the topic signal in a text (Section \ref{subsec:representation}). 

From the bigram shift results on BERT representations, we can observe that the initial layer of BERT (BERT0) and GloVe have similar performance (same as that of random embedding). Similar to our previous argument, if the initial layer of BERT encodes non-topic signals along with topic signals then we would expect that the probing model achieves a better performance (as compared to a random embedding) on a non-topic sensitive task such as bigram shift. So similar to GloVe representation, we can argue that the initial layer of BERT also primarily encodes the topic signal. But, when we look at the RoBERTa results on the bigram shift task, the initial layer achieves better performance than the GloVe baseline. This suggests that, unlike BERT, RoBERTa encodes some non-topic signals even in the initial layer. But, both BERT and RoBERTa improve their performance by using their later layers and we hypothesise that this is because both the models encode more non-topic signals in their later layers. In that case, the difference in performance between GloVe and the later layers of BERT and RoBERTa can be attributed to the encoded non-topic information that is useful to this non-topic sensitive task, for example, syntactic information. When we compare the later layer performances of BERT and RoBERTa, BERT's performance converges with that of RoBERTa and BERT's best seen score performance of 0.9447 AUC surpasses the best RoBERTa seen score of 0.9390 AUC. This suggests that even though BERT encodes less (or no) non-topic signal in its initial layers, compared to RoBERTa it encodes more non-topic signal in its later layers. Note that the drop in performance in the final layers of BERT and RoBERTa for the bigram shift probing task is likely due to the fine-tuning of the embeddings in these layers to the specific tasks that BERT and RoBERTa was trained on, namely: masked language modelling (in the case of both BERT and RoBERTa) and next sentence prediction (in the case of BERT). This pattern of performance drop is also reported in other layer-wise studies of BERT (see e.g., \cite[]{JawaharSS19}). We also observe a similar drop in performance in the last layers of BERT and RoBERTa for the general idiom token identification task and attribute the same root cause to it for that task.

The results from the bigram shift task also suggest two methods for assessing the sensitivity of a probing task to topic information. The first method is to consider the difference between a probe's performance on a task when it is trained on random embeddings versus GloVe embeddings. Tasks which have the same performance on random embeddings and GloVe embeddings are likely insensitive to the topic information (such as is the case with the bigram shift task). A corollary to this is that tasks for which there is a large difference in performance between GloVe and random embeddings are likely to be sensitive to information relating to topic. The second method to measure a task's sensitivity to the topic information is to use the difference between seen and unseen topic scores. For example, on the bigram shift task, the difference between seen and unseen topic scores is negligible for GloVe, and also for each of the layers of BERT and RoBERTa. Later in the paper, we will use both of these methods (GloVe versus random, and seen versus unseen topic scores) to assess task sensitivity to the topic information, and we will show across a range of probing tasks that these two methods are highly correlated. 

\begin{figure}
    \centering
    \includegraphics[height=4.7cm]{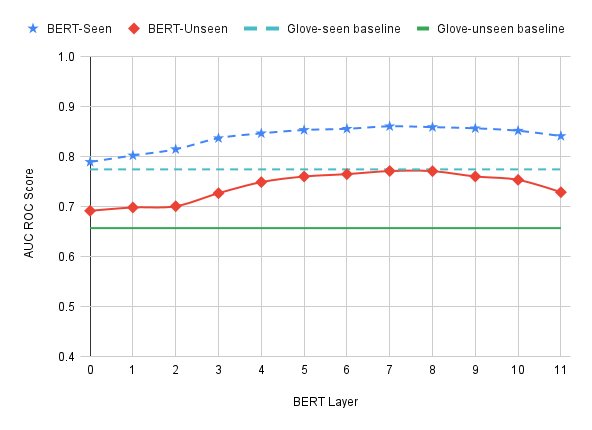}
    \includegraphics[height=4.7cm]{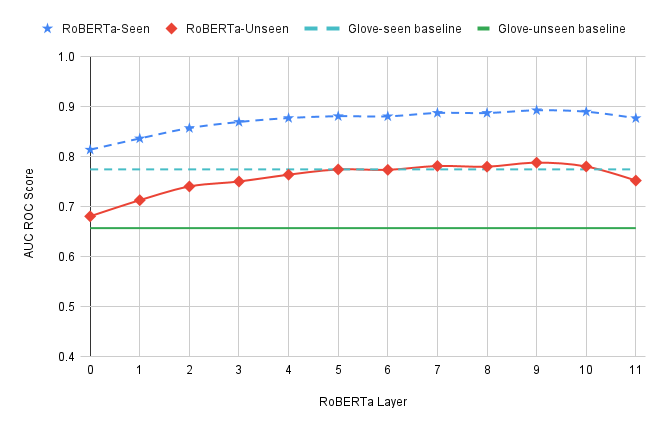}
    \vspace{0.5cm}
    \caption{Seen and Unseen AUC ROC scores from different layers of BERT and RoBERTa with GloVe baseline on General Idiom Token Identification Task}
    \label{fig:idiom-scores-layers}
\end{figure}

Switching focus to the results for the general idiom token identification task listed in Table \ref{tab:results} there are very large (as compared with the bigram shift task) performance differences between average seen and unseen topic scores on GloVe embeddings with 11.77\% difference (as compared to $<1\%$ difference on GloVe for bigram shift) and all 12 layers of BERT and RoBERTa embeddings with all differences in the range of 8.81\% to 13.34\% (as compared with a maximum difference of 2.55\% for the bigram shift task). This difference in performance between seen and unseen topics is also apparent in Figure \ref{fig:idiom-scores-layers}. This difference indicates the importance of the topic signal to the task of general idiom token identification. We also observed a large standard deviation for seen and unseen scores including for random embedding and we believe that this is an artefact of the relatively small dataset (2,566 sentences) used for the general idiom token identification task experiments.

\begin{figure}
    \centering
    \includegraphics[width=10cm]{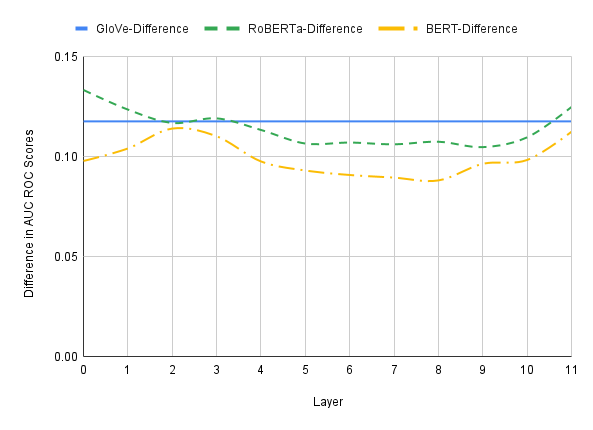}
    \vspace{0.5cm}
    \caption{Difference between seen scores and unseen scores from different layers of BERT and RoBERTa on General Idiom Token Identification Task}
    \label{fig:idiom-diff-layers}
\end{figure}

The fact that the task of general idiom token identification is sensitive to topic signal is also evidenced by the fact that GloVe performs much better than random on both seen topic and unseen topic samples. Furthermore, the difference in performance for GloVe embeddings between seen topic and unseen topic samples also reflects the ability of GloVe to encode the topic information. Also, from the results for the bigram shift task, we noted that GloVe had a similar performance to BERT0. We see a similar pattern of results here, with GloVe and BERT0 embeddings resulting in a similar performance for both the seen and unseen probing conditions. This reinforces our earlier observation that BERT0 may be primarily encoding the topic information, but also suggests that GloVe and BERT generally have a similar capacity to encode the topic signal (with BERT being slightly better at capturing this signal). In the case of RoBERTa, we observed that the initial layer encodes some non-topic signal in the bigram shift task results. But, from the results for general idiom identification on RoBERTa, we can see that the initial layer performance is similar to BERT's initial layer and GloVe. This means that, even though the first layer of RoBERTa encodes some non-topic signal, that non-topical information is not that useful for the idiom identification task. Also, as with the bigram shift task, we can observe on the general idiom token identification task an improvement in the performance of BERT and RoBERTa embeddings as we move into deeper layers, with the best BERT performance being BERT7 and the best RoBERTa performance being RoBERTa9 (on both seen and unseen conditions). This suggests that similar to our observation on the bigram shift task, the improvement in performance observed in BERT's and RoBERTa's deeper layers is attributable to non-topic-based information encoded in BERT and RoBERTa. This hypothesis is reinforced by the fact that the improvement in performance across BERT and RoBERTa layers is similar across both the seen and unseen conditions (i.e., an improvement in one layer for the seen is matched by a similar improvement for the same layer for the unseen condition). Indeed, Figure \ref{fig:idiom-diff-layers} plots the difference in BERT's and RoBERTa'a performance between seen and unseen conditions across the different layers and highlights that this difference is relatively stable and similar to the difference between GloVe for seen and unseen. It is also important to note that the difference between seen score and unseen score of all RoBERTa layers are consistently greater than that of the corresponding layers of BERT. This suggests that even though RoBERTa achieves a better performance than BERT on general idiom token identification it is more sensitive to the topic information.

To summarise, our results suggest that: (a) the bigram shift task is not sensitive to the topic information whereas the general idiom token identification task is; (b) the GloVe embeddings primarily encode topic information; (c) that initial layer embeddings of BERT (not RoBERTa) behave similarly to GloVe embeddings; (d) that later layers of BERT and RoBERTa encode non-topic information that is useful for both bigram shift and general idiom token identification (we hypothesize that this information may be syntactic in nature); (e) that both BERT and RoBERTa are sensitive to the topic information on general idiom identification. 

\section{Other Probing tasks}
\begin{figure}
    \centering
    \includegraphics[width=6cm,height=5cm]{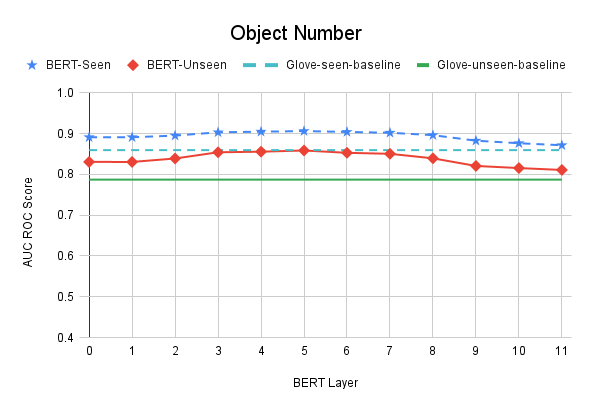}
    \includegraphics[width=6cm,height=5cm]{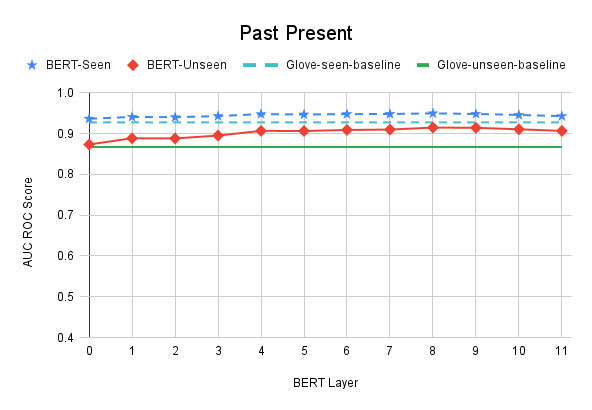}
    \includegraphics[width=6cm,height=5cm]{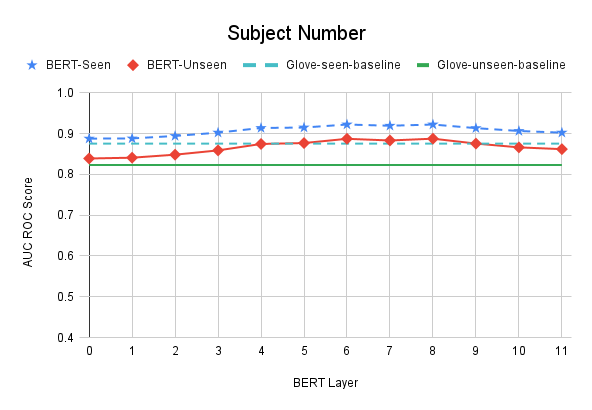}
    \includegraphics[width=6cm,height=5cm]{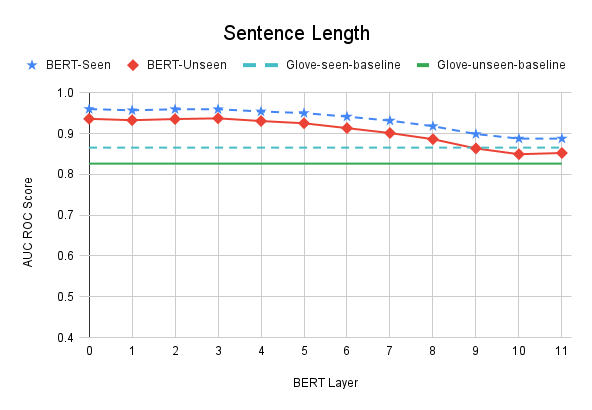}
    \includegraphics[width=6cm,height=5cm]{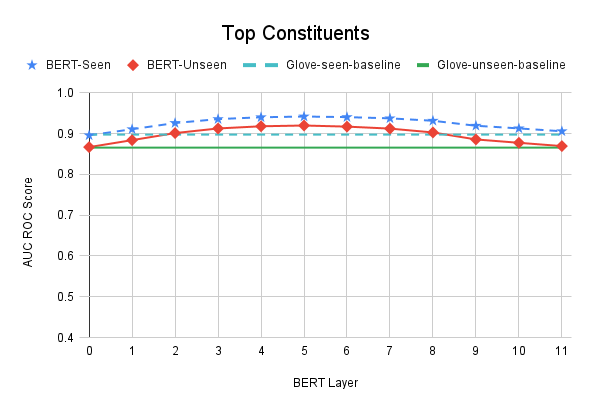}
    \includegraphics[width=6cm,height=5cm]{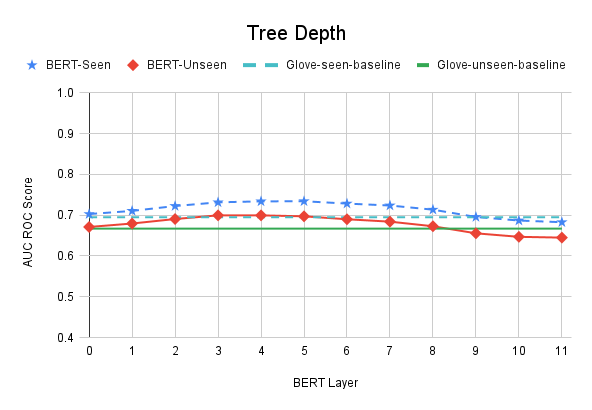}
    \includegraphics[width=6cm,height=5cm]{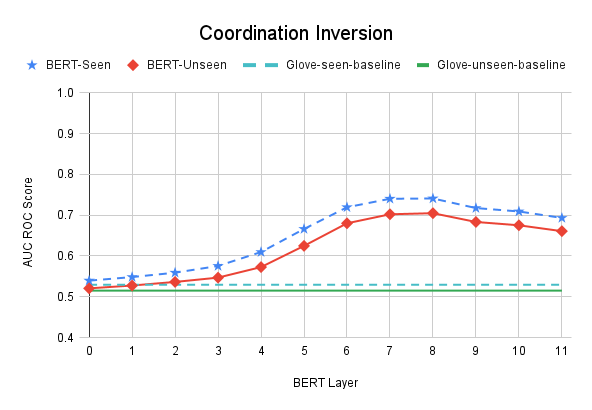}
    \includegraphics[width=6cm,height=5cm]{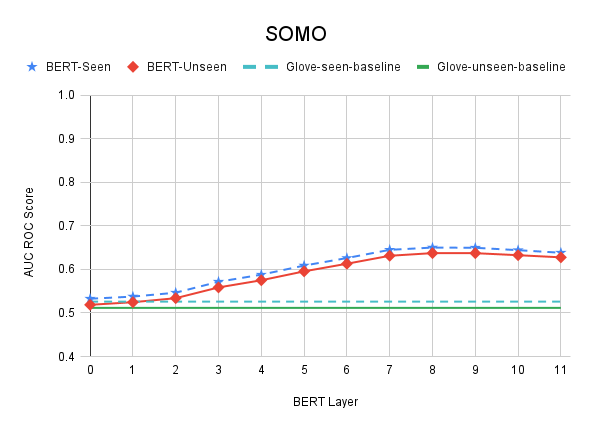}
    \vspace{0.5cm}
    \caption{Seen and Unseen AUC ROC scores from different layers of BERT with GloVe baseline on Probing Tasks}
    \label{fig:other-bert-probes}
\end{figure}

\begin{figure}
    \centering
    \includegraphics[width=6cm,height=5cm]{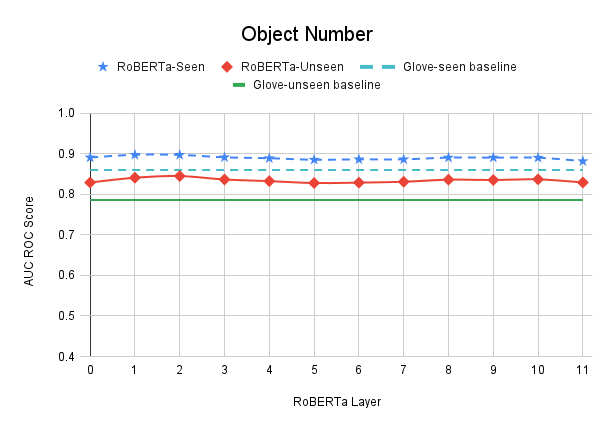}
    \includegraphics[width=6cm,height=5cm]{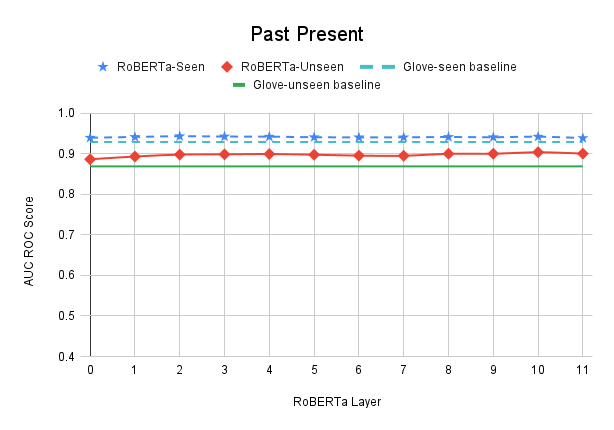}
    \includegraphics[width=6cm,height=5cm]{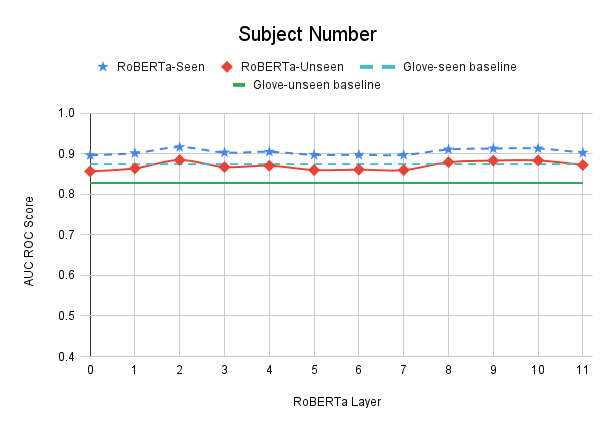}
    \includegraphics[width=6cm,height=5cm]{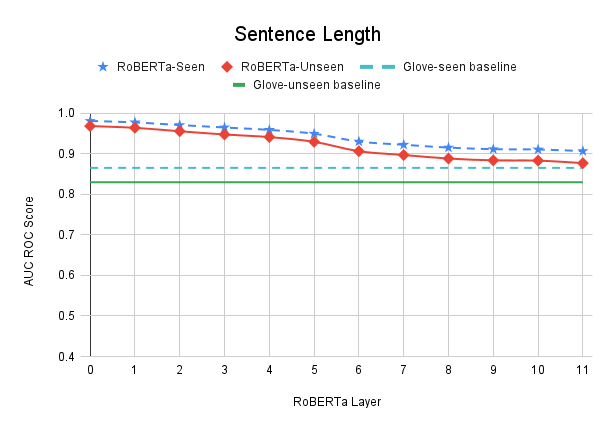}
    \includegraphics[width=6cm,height=5cm]{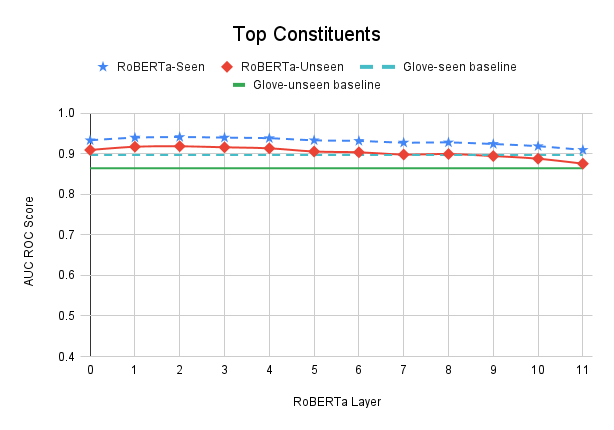}
    \includegraphics[width=6cm,height=5cm]{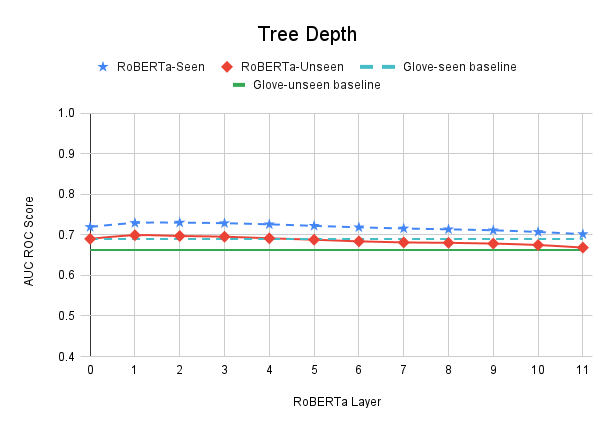}
    \includegraphics[width=6cm,height=5cm]{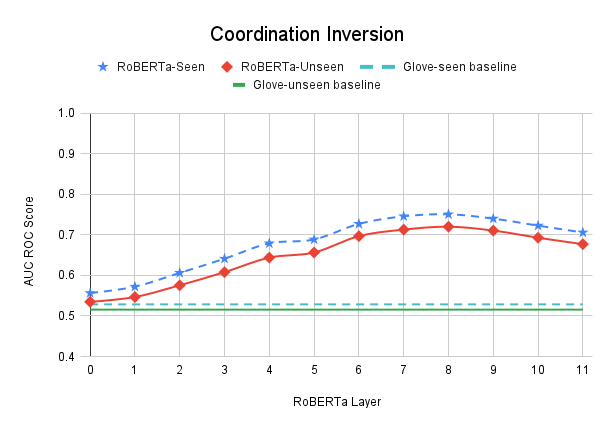}
    \includegraphics[width=6cm,height=5cm]{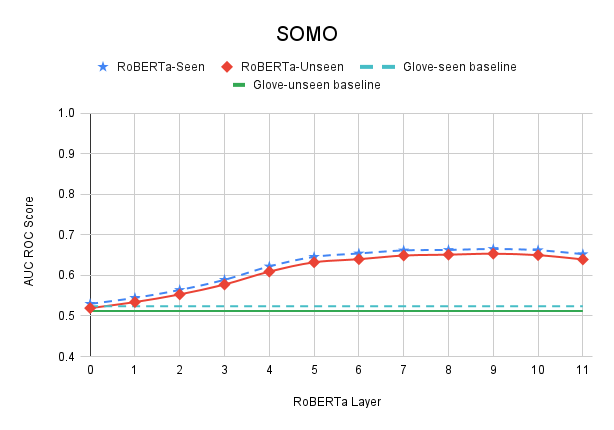}
    \vspace{0.5cm}
    \caption{Seen and Unseen AUC ROC scores from different layers of RoBERTa with GloVe baseline on Probing Tasks}
    \label{fig:other-roberta-probes}
\end{figure}

Broadening the focus beyond predicing idiomatic usage we have analysed the relationship between topic, task and the performance of probing models trained on transformer-generated embeddings across a range of standard probing tasks. We selected 8 probing tasks introduced by \citet{Conneau:2018} and the explanation and summary statistics of the dataset of each of these are shown in Table \ref{tab:other-probes-dataset}. We did a topic-aware probing on each of these 8 probing tasks and the number of tail topics obtained from different topic models on different datasets are shown in Table \ref{tab:no-topics-other-probes}. Figures \ref{fig:other-bert-probes} and \ref{fig:other-roberta-probes} show for the 8 probing tasks the performance of GloVe and the different layers of BERT and RoBERTa on each of the tasks. Looking at Figures \ref{fig:other-bert-probes} and \ref{fig:other-roberta-probes} a number of general observations can be made. First for nearly all the tasks---with the exception of Sentence Length, and Object Number (to a lesser degree)---the performance of the initial layer of BERT and GloVe is very similar in both the unseen and seen conditions. It is worth noting that in this set of probing tasks, the Sentence Length dataset is distinctive because the sentences for this task are not controlled for sentence length whereas sentences used for all the other tasks have a similar length. As pointed out by \cite{adi2017} sentence length can have a significant impact on the norm of the sentence embedding. Norms of sentence embeddings from Sentence Length datasets generated by averaging the embeddings of each word in the sentence can have a higher variance compared to other datasets and this variance may be the reason for the different behaviour observed for the Sentence Length probing task. Deviating from this pattern, the initial layer of RoBERTa achieves better performance than GloVe and the initial layer of BERT on a number of probing tasks. But for some tasks like SOMO, Coordination Inversion, and Past-Present up to some extent (and for general idiom identification task in section \ref{sec:results}), the initial layer of RoBERTa shows similar performance to that of GloVe and the initial layer of BERT. Second, in all tasks for GloVe, BERT, and RoBERTa, the performance in the seen condition is better than the unseen condition. Third, the difference in performance between the seen and unseen conditions remains relatively stable across all the layers of BERT and RoBERTa.
 
\begin{table}
   \caption{Descriptions and summary statistics of the datasets for the VNIC, Bigram shift and 8 other probing tasks}
     \begin{tabular}{@{\extracolsep{\fill}}l r r p{7cm}}
\hline \hline 
Task & No. Samples & No. Labels & Description \\
\hline					
VNIC (Idiom) & 2,566 & 2 & Predict whether a sentence contains idiomatic usage of expression from the category of verb-noun idiomatic constructions \\
Bigram Shift (BShift) & 119,998 & 2 & Predict whether any two consecutive tokens in a sentence have been inverted \\
Object Number (Obj) & 95,906 & 2 & Predict the grammatical number of the object of the main clause of a sentence \\
Past Present (PP) & 102,868 & 2 & Predict the tense of the main verb of a sentence \\
Subject Number (Subj) & 97,896 & 2 & Predict the grammatical number of the subject of the main clause of a sentence	\\
Sentence Length (SentLen) & 99,840 & 6 & Predict the length of a sentence, where sentence lengths have been binned into 6 equal-width bins \\
Top Constituents (TC) & 83,640 &	20 & Predict the syntactic top constituent of a sentence \\
Tree Depth (TreeDepth) & 50,295	& 7 & Predict the maximum depth of the parse tree of a sentence \\
Coordination Inversion (CI) & 120,004 & 2 & Predict whether two coordinated clausal conjoints in a sentence are inverted \\
SOMO & 99,986 & 2 & Predict whether a noun or a verb in a sentence has been replaced with another noun or verb \\			
     \hline \hline 
     \end{tabular}
   \label{tab:other-probes-dataset}
 \end{table}

 \begin{table}
    \centering
   \caption{Number of tail topics from 10 topic models on datasets of other 8 probing tasks}
     \begin{tabular}{@{\extracolsep{1pt}}l c c c c c c c c c c c}
     \hline \hline 
    Max number of topics & 5 & 10 & 15 & 20 & 25 & 30 & 35 & 40 & 45 & 50 \\\hline	
Object Number	&	0	&	0	&	0	&	0	&	0	&	0	&	0	&	0	&	1	&	0	\\
Past Present	&	1	&	1	&	0	&	0	&	2	&	2	&	1	&	1	&	2	&	1	\\
Subject Number	&	0	&	0	&	0	&	0	&	0	&	0	&	0	&	1	&	0	&	0	\\
Sentence Length	&	0	&	1	&	2	&	3	&	2	&	1	&	1	&	3	&	2	&	1	\\
Top Constituents	&	2	&	3	&	1	&	3	&	4	&	4	&	6	&	7	&	8	&	7	\\
Tree Depth	&	1	&	1	&	2	&	2	&	3	&	3	&	5	&	2	&	3	&	3	\\
Coordination Inversion	&	0	&	1	&	0	&	0	&	0	&	0	&	0	&	0	&	0	&	0	\\
SOMO	&	0	&	0	&	0	&	0	&	0	&	0	&	0	&	0	&	0	&	0	\\ \hline \hline 
     \end{tabular}
   \label{tab:no-topics-other-probes}
 \end{table}

For each task, we list the average seen score obtained from random embeddings, the average GloVe seen and unseen scores along with their difference, and the average seen and unseen scores and their difference from the best BERT layer and the best RoBERTa layer in Table \ref{tab:other-probes}. The best BERT layer and RoBERTa layer are the layers which obtain the best average seen AUC ROC scores, and in most of the tasks (except a very small difference in Subject Number and Tree Depth for BERT and Past Present, Tree Depth, and Object Number for RoBERTa) the best BERT layer and RoBERTa layer also gives the best average unseen AUC ROC score. For all these exceptional cases the difference between the best unseen score and the unseen score from the best layer is negligibly small (0.0003 for Subject Number and 0.0024 for Tree Depth on BERT best layers; 0.0019 for Tree Depth, 0.0014 for Past Present, and 0.0041 for Object Number on RoBERTa best layers).

Earlier we proposed two metrics derived from the topic-aware probing methodology for measuring the sensitivity of a probing task to the topic signal. The first metric is the difference between seen and unseen scores from the topic-aware probe, the larger the difference the more sensitive the task to the topic signal. The second method is based on the premise that GloVe embeddings primarily encode topic information, and so the difference between the seen scores on a task obtained from GloVe embeddings and from random embeddings can be considered as another measure of the sensitivity of a task to topic information. If both these measures are true indications of the topic sensitivity of tasks, then there should be a high correlation between them across tasks. We used the set of selected probing tasks to verify whether these measures are correlated or not. From the scores reported in Table \ref{tab:other-probes}, we calculated the correlation between these two measures of topic sensitivity from all probing tasks, i.e., (a) the difference between GloVe seen scores and Random seen scores (b) the difference between GloVe seen scores and GloVe unseen scores. We found a very high correlation coefficient, 0.80, between these measures which indicates that both these measures of task sensitivity to the topic information are consistent with each other.

 \begin{table}
   \caption{Average Seen and Unseen AUC ROC scores and their differences for GloVe and best BERT and RoBERTa layer embeddings on different probing tasks - tasks are ranked in the descending order of the difference between GloVe Seen score and GloVe Unseen score.}
    \begin{tabular}{llllllllll}
\hline \hline 
 & & Obj & PP & Subj & SentLen & TC & TreeDepth & CI & SOMO \\\hline
Random	&	Seen	&	0.5143	&	0.5136	&	0.5103	&	0.5102	&	0.5160	&	0.5047	&	0.5018	&	0.4992	\\\hline
\multirow{3}{*}{GloVe}	&	Seen	&	0.8597	&	0.9276	&	0.8755	&	0.8657	&	0.8976	&	0.6948	&	0.5295	&	0.5261	\\
	&	Unseen	&	0.7873	&	0.8672	&	0.8231	&	0.8264	&	0.8658	&	0.6670	&	0.5149	&	0.5118	\\
	&	Difference	&	0.0724	&	0.0604	&	0.0524	&	0.0394	&	0.0318	&	0.0278	&	0.0147	&	0.0144	\\\hline
\multirow{4}{*}{BERT}	&	Best Layer	&	5	&	8	&	6	&	3	&	5	&	5	&	8	&	8	\\\cline{2-10}
	&	Seen	&	0.9063	&	0.9497	&	0.9224	&	0.9599	&	0.9420	&	0.7342	&	0.7410	&	0.6501	\\
	&	Unseen	&	0.8587	&	0.9149	&	0.8873	&	0.9375	&	0.9199	&	0.6972	&	0.7049	&	0.6374	\\
	&	Difference	&	0.0476	&	0.0348	&	0.0351	&	0.0224	&	0.0221	&	0.0370	&	0.0361	&	0.0127	\\\hline
\multirow{4}{*}{RoBERTa}	&	Best Layer	&	1	&	2	&	2	&	0	&	2	&	2	&	8	&	9	\\\cline{2-10}
	&	Seen	&	0.8974	&	0.9433	&	0.9164	&	0.9809	&	0.9412	&	0.7301	&	0.7507	&	0.6651	\\
	&	Unseen	&	0.8409	&	0.8979	&	0.8846	&	0.9682	&	0.9185	&	0.6972	&	0.7201	&	0.6535	\\
	&	Difference	&	0.0565	&	0.0453	&	0.0318	&	0.0127	&	0.0227	&	0.0329	&	0.0306	&	0.0116	\\
     \hline \hline 
     \end{tabular}
   \label{tab:other-probes}
 \end{table}

\begin{figure}
    \centering
    \includegraphics[width=12cm]{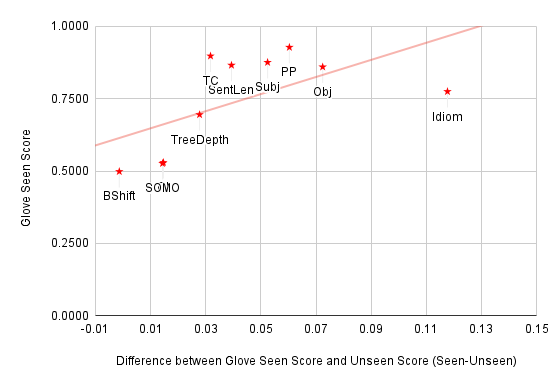}
    \vspace{0.5cm}
    \caption{GloVe Seen Score versus GloVe Score Difference (Task Topic Sensitivity) for each probing task (Note that scores of SOMO and CI are very similar and therefore both of them are overlapping in the plot)}
    \label{fig:GloVe-Seen-vs-GloVe-Diff}
\end{figure}

\begin{figure}
    \centering
    \includegraphics[height=10cm,width=12cm]{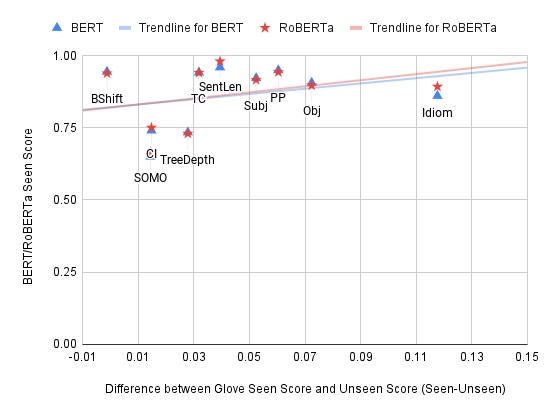}
    \vspace{0.5cm}
    \caption{BERT and RoBERTa Seen Score versus GloVe Score Difference (Task Topic Sensitivity) for each probing task}
    \label{fig:BERT-Seen-vs-GloVe-Diff}
\end{figure}

By taking the difference between GloVe seen performance and GloVe unseen performance as the measure of topic sensitivity, we generated two scatter plots to check how the performance of GloVe and Transformer-based neural language models (BERT and RoBERTa) varies across different tasks with different topic sensitivity. Figure \ref{fig:GloVe-Seen-vs-GloVe-Diff} plots for each task the GloVe seen performance versus the difference between GloVe seen and unseen and Figure \ref{fig:BERT-Seen-vs-GloVe-Diff} plots for each task the best performance by BERT and RoBERTa (for any layer) versus the difference between GloVe seen and unseen. In these scatter plots, we can see a general trend that as sensitivity to topic increases (i.e., the difference between GloVe seen and unseen gets larger and we move to the right on the x-axis) there is a tendency for the performance of GloVe and Transformer based neural language models (BERT and RoBERTa) to increase. This suggests that the less sensitive a probing task is to topic information the more difficult the task is for Tranformer-based neural language models (BERT and RoBERTa). We also calculated the correlation coefficients between performances of neural language models (seen scores) and topic sensitivity scores (difference between GloVe seen and unseen score), and we found a coefficient of 0.2829 for BERT and 0.3345 for RoBERTa. This suggests that neural language models rely on topical information to solve different tasks.

When we compare the BERT and RoBERTa neural language models we observe slightly more topic reliance for RoBERTa than BERT. This difference is evident in Figure \ref{fig:BERT-Seen-vs-GloVe-Diff} (steeper trend line for RoBERTa) and in a slightly higher value of correlation coefficient for RoBERTa than BERT (0.3345 vs 0.2829). Our results indicate that newer and generally better neural language models, such as RoBERTa, are more reliant on topic information as compared with BERT. We will return to this point in our conclusions. 

\section{Conclusions}\label{sec:conclusions}
We proposed a topic-aware probing method to measure the role of the topic signal in distributed representations and validated this method using baseline representations (GloVe and random) and a baseline task (bigram shift). The results of our analysis on the bigram shift probing task supported our hypothesis that GloVe embeddings primarily encode topic information, and furthermore suggested that the initial layer of BERT also primarily encodes topic information and that later layers of BERT and all layers of RoBERTa encode non-topic information (the observation that this information was non-topic related is based on the fact that this information was useful for the non-topic sensitive probing task of bigram shift). 

With respect to the task of general idiom token identification, previous research has pointed to the importance of topic information for identifying idiomatic usage \cite[]{Feldman:2013,Peng:2014}. However, work such as by \citet{Fazly:2009} has highlighted the importance of non-topic information for idiom token identification and the results of \citet{Salton:2016} suggested that by using distributed embeddings it is possible to create a general idiom token identification model (i.e., a model that works across multiple expressions within a category) without requiring large amounts of topic information. The results of our topic-aware probing experiments confirm the importance of the topic signal to the task of general idiom token identification. One potential reason for the topic signal being important for general idiomatic token identification systems is that distinctions between topics may align with seen versus unseen expressions. Our analysis of the distribution of expressions across topics (see section \ref{subsec:probing-experiment}) suggested that for many expressions the sample sentences containing the expression tend to cluster within a topic, particularly the literal uses of an expression. Consequently, training a probe on one topic and testing it on other topics is similar to training on one set of expressions and testing on another set of expressions. Indeed, \cite{Nedumpozhimana:2022} reports an expression-based analysis of idiom token identification that includes a seen versus unseen experiment. The results of that experiment also reported a drop in performance, in this case from seen expressions to unseen expressions rather than seen versus unseen topics. The fact that the distinction between seen and unseen topics and seen and unseen expressions overlaps is not surprising. However, the overlap does not mean that the phenomena are identical. For example, whereas the expression-based distinction is solely based on the words within an expression, the definition of topic in our experimentation considers all words in a sentence. The importance of considering both the information within an expression and the surrounding context for idiom token identification has been demonstrated by  \mbox{\citet{nedumpozhimana2021finding}} who show that BERT not only relies on information within the expression but also in the surrounding context.

Switching to our primary question of the extent to which Transformer-based language models rely on word order/syntactic information versus word co-occurrence/topic information, the results of our topic-aware probing experiments suggest that BERT and RoBERTa encode both topic and non-topic information. This is indicated by the fact that across all the probing tasks the embeddings generated by BERT and RoBERTa's middle layers result in a higher performance than  GloVe, see Figure \ref{fig:other-bert-probes}. These results are in line with various layer-wise studies on BERT in the literature, such as by \citet{JawaharSS19}, which suggest that the syntactic features of a sentence are encoded in the middle layers of BERT. However, despite the fact that BERT can capture useful non-topic information our analysis suggests that in general BERT (and RoBERTa) primarily rely on the topic information. Furthermore, our analysis of BERT and RoBERTa's performance across a set of standard probing tasks suggests that tasks that are relatively insensitive to the topic information are also tasks that are relatively difficult for BERT and RoBERTa. These observations agree with the findings of \cite{pham-etal-2021-order}, in which they observe that most of the BERT-based models behave similarly to bag-of-word models on GLUE tasks. They also agree with \cite{sinha-etal-2021-masked} who argue that the success of pre-trained language models on many tasks is primarily based on their ability to encode distributional information.

From our experiments, we observed that the RoBERTa model is more reliant on the topic information than the BERT model. One of the most notable differences between BERT and RoBERTa is in the pre-training objectives used for the two models. BERT is trained on the masked language model and the next sentence prediction objectives, whereas the RoBERTa model excludes the next sentence prediction objective. \cite{mickus-etal-2020-mean} argue that the use of a next-sentence prediction objective adulterates the distributional nature of the semantics learned by BERT. Building on this argument the removal of the next sentence prediction from the pre-training objective for RoBERTa may result in the RoBERTa model being more focused towards distributional semantics, and this may explain the relatively stronger reliance of RoBERTa on topic compared to BERT.

The broader implications of our findings for NLP is that the performance of Transformer-based systems on NLP tasks can be improved by incorporating more word order or syntactic information into these language models. \cite{Wang-structbert-2020} is an example of recent work that attempts to do this. They show that adding word-order and sentence-order learning objectives into BERT pretraining could lead to improved performance on language processing tasks. Also, \cite{pham-etal-2021-order} show that fine-tuning on a word-order sensitive task prior to fine-tuning on a downstream task increases a language model's sensitivity to word order on the downstream task. Another potential way forward in this direction is to explicitly integrate syntactic information into the language modelling architecture, for example by using methods like Recursive Neural Network in which the representation of a sentence is composed of representations of words by applying compositions recursively through the parsed tree \citep{socher-etal-2013-recursive}. Such approaches utilize compositional semantics which can be helpful to capture more non-distributional semantics in language models.

There are a number of limitations of our analysis that should be noted. The first is that the analysis is based on two neural language models BERT and RoBERTa, both are based on Transformer encoder architecture. Recent neural language models like GPT, Llama, etc. are based on Transformer decoder architecture. It is natural to ask whether we can extend our study to neural language models with Transformer decoder architecture and whether the observations from our experiment can be applicable to those models. Also, our analyses are based only on English datasets which have relatively fixed word order. It would be interesting to examine whether the observations we have drawn from our experiments are limited to the English language or whether the reliance of Transformer-based language models on topic as strong for other languages that have a relatively more flexible word order and richer morphology. In future work, we will extend our experiments by using different neural language models based on the Transformer decoder architecture and different datasets like PARSEME \citep{savary-etal-2017-parseme, ramisch-etal-2018-edition, ramisch-etal-2020-edition}, MAGPIE \citep{haagsma-etal-2020-magpie}, etc., which contains both English and Non-English data to address these questions.

Our findings regarding the topic reliance of neural language models have implications for NLP tasks such as machine translation and question answering. For example, the findings of \citet{amponsah-kaakyire-etal-2022-explaining} indicate that part of BERT's performance on the task of identifying translationese is due to topic differences learned by BERT. So, another direction for future work is to use topic-aware probing to investigate the topic reliance of neural language models on identifying translationese. 

\bibliographystyle{unsrtnat}
\bibliography{reference}  


\end{document}